\documentclass[10pt,journal,compsoc]{IEEEtran}
 \usepackage{color}
 \usepackage{hyperref}
 \usepackage{multirow}
\usepackage{cite}

\ifCLASSINFOpdf
   \usepackage[pdftex]{graphicx}
\else
\fi
\usepackage{comment}

\usepackage{amsmath,amssymb}
\DeclareMathOperator*{\argmax}{arg\,max}

\ifCLASSOPTIONcompsoc
  \usepackage[caption=false,font=normalsize,labelfont=sf,textfont=sf]{subfig}
\else
  \usepackage[caption=false,font=footnotesize]{subfig}
\fi

\usepackage{float}
\usepackage{url}

\usepackage{xcolor}
 
\begin{document}
%
\title{Continual Learning Using Multi-view Task Conditional Neural Networks}
%
%
%

\author{Honglin Li,
        Payam Barnaghi,~\IEEEmembership{Senior Member IEEE},
        Shirin Enshaeifar,~\IEEEmembership{Member IEEE},
        Frieder Ganz 
\thanks{H. Li, P. Barnaghi and S. Enshaeifar are with the Centre for Vision, Speesh and Signal Processing (CVSSP) at the University of Surrey and Care Research and Technology Centre at the UK Dementia Research Institute (UK DRI). email: \{h.li,p.barnaghi,s.enshaeifar\}@surrey.ac.uk}
\thanks{P. Barnaghi is also with the Department of Brain Sciences, Imperial College London. email: {p.barnaghi}@imperial.ac.uk}
\thanks{F. Ganz is with the Adobe, Germany. email: {ganz}@adobe.com}
\thanks{* H. Li and P. Barnaghi have equally contributed to this work.}}

%
%

\markboth{Journal of \LaTeX\ Class Files,~Vol.~14, No.~8, August~2015}%
{Shell \MakeLowercase{\textit{et al.}}: Bare Demo of IEEEtran.cls for IEEE Journals}
%



\IEEEtitleabstractindextext{%
\begin{abstract}
Conventional deep learning models have limited capacity in learning multiple tasks sequentially. The issue of forgetting the previously learned tasks in continual learning is known as catastrophic forgetting or interference. When the input data or the goal of learning changes, a continual model will learn and adapt to the new status. However, the model will not remember or recognise any revisits to the previous states. This causes performance reduction and re-training curves in dealing with periodic or irregularly reoccurring changes in the data or goals. The changes in goals of learning or data are referred to as new tasks in a continual learning model. Most of the continual learning methods have a task-known setup in which the task identities are known in advance to the learning model. We propose a Multi-view Task Conditional Neural Network (Mv-TCNN) that does not require to know the reoccurring tasks in advance. Mv-TCNN provides a novel approach compared to the state-of-the-art models to handle the catastrophic forgetting problem, which require advance information regarding the tasks and also offers better adaptability and performance in learning reoccurring tasks. We evaluate our model on standard datasets using MNIST, CIFAR10, CIFAR100, and also a real-world numerical time-series dataset that we have collected in a remote healthcare monitoring study (i.e. TIHM dataset). The proposed model outperforms the state-of-the-art solutions in continual learning and adapting to new tasks that are not defined in advance.
\end{abstract}

}

\maketitle



%
\IEEEpeerreviewmaketitle

\section{Introduction}
\IEEEPARstart{T}he human brain can adapt and learn new knowledge in response to the changing environments. We can continually learn different tasks while retaining previously learned variations of the same or similar phenomenon and give different reactions under different contexts. 
Neurophysiology research has found that our neurons are task-independent \cite{asaad2000task}. Under different context, the neurons are fired selectively with respect to the stimulus. In contrast, most of the machine learning models, in a scalable way, are not capable of adapting to changing environments quickly and automatically using neurons corresponding to different tasks. As a consequence, these machine learning models tend to forget the previously learned task after learning a new task. This scenario is known as catastrophic forgetting or interference in machine learning \cite{mccloskey1989catastrophic}.

Catastrophic interference problem in machine learning is one of the hurdles to implement a general artificial intelligence learning system without constructing a set of models each dedicated to a specific task or different variations and situations in the data \cite{legg2007universal}. Unable to learn several tasks, the model should be trained with all the possible scenarios in advance. This requirement is intractable in practice and is not inline with the lifelong learning goal in continual learning models \cite{thrun1995lifelong}. 

Continual machine learning algorithms change over time and adapt their parameters to the changes in data or the learning goal. We refer to the learning goal or a specific part of the data with a learning goal as a \textit{task}. The learning models are not often equipped with solutions to quickly adapt to the situations which they have seen before if their parameters have significantly changed over time by continual learning. Here we use an example to illustrate the forgetting problem. We train a neural network for two different tasks sequentially. After being trained for each task, the model is represented by parameters $\theta_0,\theta_1,\theta_2$ respectively, where the $\theta_0$ is the randomly initialised weights, $\theta_1$ is the weights after learning task 1, $\theta_2$ is the weights after learning task 2. We use the linear path analysis \cite{goodfellow2014qualitatively} to visualise the loss surface. We define $\theta = \theta_0 + \alpha(\theta_1 - \theta_0) + \beta(\theta_2 - \theta_1)$.

\begin{figure}
    \centering
    \includegraphics[width=\linewidth]{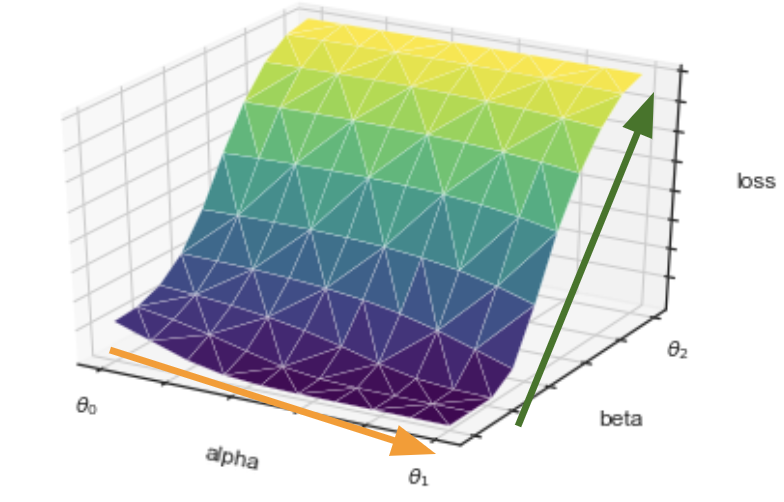}
    \caption{Loss surface for the first task while learning two tasks sequentially. $\theta_0$ represents the initial parameters, $\theta_1$ and $\theta_2$ are the optimal solutions for the first and second task respectively. While learning the first task, the loss decrease along the orange arrow. While learning the second task, the loss for the first task starts to increase along the green arrow.}
    \label{fig:intro_loss_surface}
\end{figure}

As shown in Figure \ref{fig:intro_loss_surface}, while the model learns the first task, the parameters change from $\theta_0$ to $\theta_1$, and the loss for the first task becomes smaller along with the orange arrow. When the model learns the second task, the parameters change from $\theta_1$ to $\theta_2$, and the loss for the first task increases significantly along with the green arrow.

A real-world example of this problem is a challenge that we have faced in our remote healthcare monitoring study \cite{enshaeifar2018internet}. We have developed a digital platform and a set of machine learning algorithms to perform risk analysis and provide alerts for early interventions in a use-case scenario to support people affected by dementia. In our user group with an in-home monitoring environment, the distribution of data is periodical (due to time of the day, seasonal and environmental effects). In some cases, the data and conditions sporadically change and repeat due to variations in participants' health conditions. When we use continual and adaptive learning to analyse this data, we need to update the models according to these changes. However, we face a problem that models do not preserve earlier learned tasks when they reoccur. There are two potential solutions to this problem, either maintaining multiple models to respond to different situations or developing models that inherently adapt to the changes and preserve the previously learned tasks as well. Maintaining several models for changing tasks also faces another challenge to detect when a change has occurred and being able to identify if the same or a similar task has previously been observed. 

A variety of continual learning methods have been proposed to solve the problems mentioned above. Memory-based approaches replay the trained samples to solve the forgetting problem while learning a new task \cite{shin2017continual}. Regularisation methods reduce the representational overlap of different tasks \cite{lee2017overcoming,kirkpatrick2017overcoming,zeng2018continuous}. Dynamic network approaches assign extra neuron resources to new tasks \cite{yoon2017lifelong}. Some of the existing solutions need to know all the task changes in advance, or this information is given manually to the model throughout the learning. Some other works \cite{li2019continual, lee2020neural} try to detect the in-task and out-task samples to inference the task identity.

In this work, we propose a novel method to overcome the catastrophic forgetting problem. The proposed method combines the multi-view \cite{sun2013survey} techniques, probabilistic neural networks \cite{specht1990probabilistic} and conventional neural networks \cite{haykin1994neural} to produce the task conditional probabilities. Hence we name our approach as Multi-view Task Conditional Neural Network (Mv-TCNN). Our model contains multiple experts, each of which is responsible for a task. 

We highlight the key contributions of this work as follows.
\begin{enumerate}
    \item We propose a continual learning model combined with a multi-view method.
    \item We propose a novel probabilistic layer inspired by probabilistic neural networks. However, our model does not need to perform Monte Carlo sampling \cite{li2019continual} which is resource extensive, or it does not require an extra generative model in each expert \cite{lee2020neural} to estimate the confidence to get the task identity. 
    \item Our model can detect the task changes automatically.
    \item The proposed model has been evaluated based on several benchmark experiments of continual learning, including MNIST, CIFAR 10 and CIFAR100. We also show that our model can be applied to a real-world numerical time-series dataset in the healthcare domain (i.e. TIHM dataset).
\end{enumerate} 

\section{Different Scenarios in Continual Learning Settings}\label{sec:reason}
Kortge \textit{et. al} \cite{Kortge1990EpisodicMI} state that the interference problem in neural networks is due to the back-propagation rule. This idea is studied by several groups, including Kirkpatrick \textit{et. al} and Lee \textit{et. al} \cite{kirkpatrick2017overcoming,lee2017overcoming} have studied this idea.. They state that the reason for interference is because the parameter space adapts to a new task rapidly and then comprises the previous task but with lower accuracy in responding to the earlier learned task. While learning two tasks sequentially, the model pays attention to the current task. In this case, the parameters change significantly after learning a new task, and if the model is given the earlier task again, it would not respond well until it re-learns it again. French \textit{et. al} \cite{french1991using,french1992semi,french1999catastrophic} argue that the problem is due to the overlap in the internal representation of different tasks. Goodfellow \textit{et. al} \cite{goodfellow2013empirical} have also investigated this idea and shown that adding dropouts \cite{srivastava2014dropout} can mitigate catastrophic interference. Similarly, Masse \textit{et. al} \cite{masse2018alleviating} demonstrate that by deactivating a portion of the neurons before training new tasks, a model can address the catastrophic interference.

Based on the existing studies reported in  \cite{zenke2017continual,masse2018alleviating,nguyen2017variational,serra2018overcoming}, the task information is one of the main causes of catastrophic interference. We explore different scenarios of informing the model about the changing task information: 

\textit{S1}: The task is unknown to the model all the time; \textit{S2}: The model do not need to be informed of the task information at the testing stage \cite{lee2017overcoming,kirkpatrick2017overcoming}. However, it cannot detect the changes automatically at the training stage; \textit{S3}: The task information is known at both the training and testing stages \cite{zenke2017continual,nguyen2017variational,serra2018overcoming,masse2018alleviating}. The model needs to be told which neurons should be activated during the testing stage; \textit{S4}: The model knows what task it is about to perform before the training starts and knows the task changes during the test and run-time \cite{hetherington1993catastrophic}.

\begin{figure}[t]
    \centering
    \includegraphics[width=\linewidth]{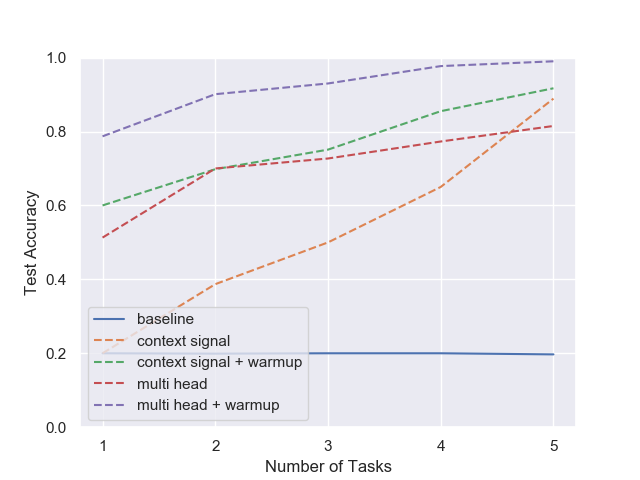}
    \caption{Test accuracy with providing different task information in advance.}
    \label{fig:intro_experiment}
\end{figure}

Figure \ref{fig:intro_experiment} demonstrates how the task information affects the results. The scenarios \textit{S1} and \textit{S2} are shown as baseline in Figure \ref{fig:intro_experiment}. In the scenario \textit{S3}, there are many different ways to inform the model about the task identities. Here we use context signal \cite{mirza2014conditional,masse2018alleviating} and multi-head approach \cite{zenke2017continual}. The context signal is to add the task identity along with the samples in the input layer. The multi-head is to mask the output layer to make the model only response to the current task. The scenario \textit{S4} is named as warmup in Figure \ref{fig:intro_experiment}. Warmup allows the model to preserve a small set of samples drawn from the tasks to be learned. Overall, the positive effect of the task information shown in Figure \ref{fig:intro_experiment}. The performance increases in the following order: i) Baseline (no task information), ii) Context signal, iii) Multi-head, iv) Context + warmup, v) Multi-head + warmup.

Since the multi-head approach tells the model in advance which task it is about to deal with, the model can determine approximate parameters, despite the model not being aware of the task beforehand. This is why the multi-head approach provides a higher overall accuracy after learning the first task.

As shown in Figure \ref{fig:intro_experiment}, a model can address the catastrophic interference problem by using advance information about the changes. The multi-head and warmup solutions explicitly provide all the task information to the model, and this allows them to learn new tasks without forgetting the previous ones. Overall, the more advance information is provided regarding a task that a model is about to encounter, the more effective the model will be in adapting to the new task.

Based on the above, one can see that the task information is important to address the forgetting problem in continual learning. Informing the task identity can be regarded as maximising the likelihood of $P(t=k|x)$, where $x$ is the samples and $t=k$ is the task information. To infer the task information, Li \textit{et. al} \cite{li2019continual} leverage the uncertainty to get the task information at the prediction stage. However, their method cannot detect the changes automatically. Farquhar \textit{et. al} \cite{farquhar2018towards} suggest that mutual information can be used to identify the changes. However, calculating mutual information can become intractable in large-scale scenarios  \cite{li2019continual}. Lee \textit{et. al} \cite{lee2020neural} proposed a non-parametric model named Continual Neural DirichletProcess  Mixture  (CN-DPM), which can detect the changes in the training state and without being informed of the task identity in the testing state. In the CN-DPM, there are many different experts cooperate with each other to learn different tasks. Each of the experts contains a generative model to estimate the confidence, and a discriminative model to do the classification. In our work, we develop an expandable model in the similar manner, but mainly different from the CN-DPM in the following aspects. Only one expert takes responsibility for one task in our work. In CN-DPM, there may be multiple experts are responsible for the same tasks. We do not need an extra generative model to produce the confidence measure. Our model combines the proposed probabilistic neural layer and multi-view functions to estimate the confidence measure. 

\section{Related work}

There are different approaches to address the forgetting problem in continual learning. Parisi \textit{et. al} \cite{parisi2019continual} categorise these approaches into three groups: Regularisation, Memory Replay and Dynamic Network approaches. 

The regularisation approaches find the overlap of the parameter space between different tasks. One of the popular algorithms in this group is Elastic Weight Consolidation (EWC) \cite{kirkpatrick2017overcoming}. EWC avoids significantly changing the parameters that are important to a learned task. It assumes the weights have Gaussian distributions and approximates the posterior distribution of the weights by the Laplace approximation. A similar idea is used in Incremental Moment Matching (IMM) \cite{lee2017overcoming}. IMM finds the overlap of the parameter distributions by smoothing the loss surface of the tasks. Zeng \textit{et. al} \cite{zeng2018continuous} address the forgetting problem by allowing the weights to change within the same subspace of the previously learned task. Li \textit{et. al} \cite{li2018learning} address the problem by using the knowledge distillation \cite{hinton2015distilling}. They enforce the prediction of the learned tasks to be similar to the new tasks \cite{parisi2019continual}. However, these models require advance knowledge of the training tasks and the task changes. 
.

Memory Replay methods mainly focus on interleaving the trained samples with the new tasks. A pseudo-rehearsal mechanism \cite{robins1995catastrophic} is proposed to reduce the memory requirement to store the training samples for each task. In a pseudo-rehearsal, instead of explicitly storing the entire training samples, the training samples of previously learned tasks are drawn from a probabilistic distribution model. Shin \textit{et. al} \cite{shin2017continual} propose an architecture consisting of a deep generative model and a task solver. Similarly, Kamra \textit{et. al} \cite{kamra2017deep} use a variational autoencoder to generate the previously trained samples. However, this group of models are complex to train, and in real-world cases, the sampling methods do not offer an efficient solution for sporadic and rare events. These models also often require advance knowledge of the change occurring.

Dynamic Networks allocate new neurons to new tasks. Yoon \textit{et. al} \cite{yoon2017lifelong} propose Dynamic Expandable Networks (DEN) to learn new tasks with new parameters continuously. Similarly, Serra \textit{et. al} \cite{serra2018overcoming} also allocate new parameters to learn new tasks. However, this group of models require the task information to be given to the model explicitly. In other words, the model knows in advance, which neurons should be activated to perform each test task. To identify which neurons or experts should be used during the test state, Aljundi \textit{et. al} \cite{aljundi2017expert} and Lee \textit{et. al} \cite{lee2020neural} leverage generative models, to achieve the task-free continual learning, which do not need the task identities in the test phase. 

\section{Multi-view Task Conditional Neural Network}
In the Mv-TCNN, we augment the training data by the multi-view functions and then train an expert for each new task. During the inference phase, we augment the test data by the same multi-view functions and give a prediction based on the task likelihood and classification results. Each expert is independent of others and has the same behaviour during the training and inference phases. In this section, we illustrate our model by taking one expert as an example. 
\subsection{Estimate the task likelihood} \label{sec:method_pnn}

\begin{figure}[t]
    \centering
    \includegraphics[width=\linewidth]{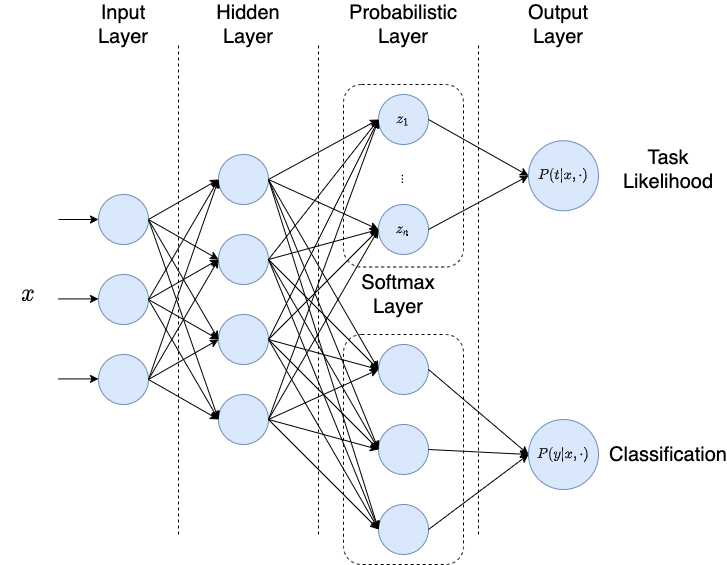}
    \caption{The architecture of an expert in the proposed model. The probabilistic layer contains multiple data patterns which can be learned by maximising the task likelihood.}
    \label{fig:model_architecture}
\end{figure}

In this work, we refer to the task likelihood as $P(t=k|x)$ to represent the sample $x$ from the $k_{th}$ task. To estimate the task likelihood efficiently, we add a probabilistic layer into the expert, as shown in \ref{fig:model_architecture}. Different from the conventional neural networks, this model contains two output heads referring to the task likelihood head and classification head. The classification head is connected to a softmax layer, which is similar to a conventional neural network. The task likelihood head is connected to the probabilistic layer. 

Different from the conventional fully-connected networks, the nodes in the probabilistic layer do not perform multiplication. We refer to the parameters of the nodes as the kernels of the probabilistic layers. The $i_{th}$ nodes in the probabilistic layer perform a potential function \cite{serpen1997performance} shown in Equation (\ref{eq:potential_function}).

\begin{equation}
\label{eq:potential_function}
    K(f(\textbf{x}), \textbf{z}_{i}) = \exp[-\frac{1}{2}(f(\textbf{x}) - \textbf{z}_{i})^T \Sigma^{-1}(f(\textbf{x})-\textbf{z}_{i})]
\end{equation}

where the $f(\cdot)$ is the function of hidden layers parameterised by $\theta_S$, $z_i$ is the kernel in the $i_{th}$ node, $\Sigma$ is the covariance matrix of $z_{1:n}$, where $n$ is the number of the nodes in the probabilistic layer. Calculating the covariance matrix could be intractable due to the high dimension of $z_{1:n}$. We assume the nodes in the probabilistic layers are independent of each other, and the diagonal values in the covariance matrix are pre-defined constants.

In the probabilistic layer, each node estimates the similarity between the input and the existing kernel in the node. This similarity is measured by using the potential function. Hence the kernels can be regarded as the data patterns of the output of the last hidden layer. Overall, the probabilistic layer can estimate the probability density of the samples during the training phase. In the inference phase, the probabilistic layer compares the test data to the kernels referring to the data patterns of the learned samples.

The output of the probabilistic layer is an $n$ dimensional vector. The elements in the vector can be regarded as the similarity of the input to all the data patterns in this task. The summation of these $n$ similarities can be viewed as the task likelihood. To estimate the task likelihood, we perform a normalised summation shown in Equation (\ref{eq:summation}).

\begin{equation}
\label{eq:summation}
    P(t|x,\cdot) = \frac{\sum_{i=1}^n K(f(\textbf{x}), \textbf{z}_{i})}{\sum\limits_{i=1}^n K(f(\textbf{x}), \textbf{z}_{i}) + 1 - \max\limits_{\small{j = 1\dots n}} \{K(f(\textbf{x}), \textbf{z}_{j})\}}
\end{equation}

While training the new expert, we jointly maximise the task likelihood and the classification likelihood shown in Equation (\ref{eq:loss_function}).

\begin{equation}
\label{eq:loss_function}
\begin{split}
    L(\theta_S, \theta_D,\theta_T) =& \sum_{c=1}^C y^{(c)}\log P(y^{(c)}|x^{(c)}, \theta_D, \theta_S)  \\
    &+ \lambda * \log P(t=k|x, \theta_T, \theta_S)
\end{split}
\end{equation}

Where ${x, y}$ is the training sample set of task $k$, $C$ is the number of classes in the task $k$, $\theta_S$ is the parameters in the hidden layers shared by classification and probabilistic layer, $\theta_D$ and $\theta_T$ are the parameters in the softmax and pattern layers respectively, $\lambda$ is a hyper-parameter parameter to weight the task likelihood loss,$P(t=k|x, \theta_T, \theta_S)$ is estimated by Equation (\ref{eq:summation}).

\subsection{Multi-view functions}
\begin{figure}
    \centering
    \includegraphics[width=\linewidth]{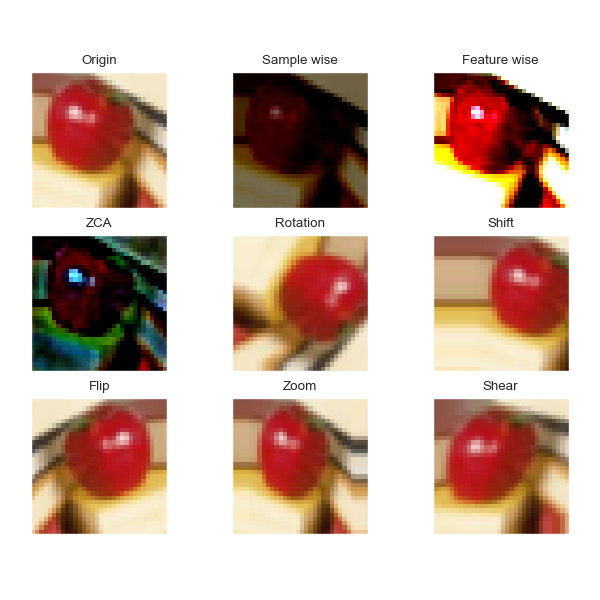}
    \caption{Different view functions applied to a sample input.}
    \label{fig:multi_view}
\end{figure}

To produce the task likelihood more accurately, we introduce the multi-view functions. Before the training starts, we pre-define a function set $G = \{g_1(\cdot), g_2(\cdot), \dots, g_m(\cdot) \}$ to augment the training data for each expert. In this work, we define nine types of functions to get a different view of the inputs. Figure \ref{fig:multi_view} shows different ways of processing the inputs. The sample wise including two functions referring to centralisation and std normalisation. The centralisation sets the mean of the input to zero. The std normalisation divides the input by its standard deviation. The feature-wise includes centralisation and std normalisation functions. But they are applied to the feature of the whole dataset. Since the samples from each task may have different means and standard deviations, different experts have their own feature-wise functions. The rotate function rotates rotate the inputs from 0 to 270 degree randomly. The shift function shifts the inputs horizontally or vertically. The flip function flips the inputs horizontally or vertically. The shear function distorts the input. The ZCA performs the Zero Component Analysis (ZCA) \cite{pal2016preprocessing}, shown in Equation (\ref{eq:zca}), 

Where $U$ is the eigenvector matrix, $S$ is the Eigenvalue matrix obtained form the singular value decomposition of the covariance matrix, $\epsilon$ is the whitening coefficient.

\begin{equation}
\label{eq:zca}
    X_{zca} = U.diag(1/\sqrt{diag(S) + \epsilon}).U^T.X^`.
\end{equation}

\subsection{Training and Inference}
\textbf{Training}: When a brand new task is detected, we train a new expert. We first augment the input $x$ by the function set $Mv$, then train the expert according to the Equation (\ref{eq:loss_function}). 

\textbf{Inference}: After training all $K$ experts, we augment the test data by the same multi-view functions used in each expert and estimate the $P(y|x)$ by Equation (\ref{eq:inference})

\begin{equation}
\label{eq:inference}
\begin{split}
        P(y|x) & \approx \frac{1}{m}\sum_{v=1}^m P(y_v|g_v(x)) \\
         \approx & \frac{1}{m}\sum_{k=1}^K \sum_{v=1}^m P(y_v|g_v(x), t=k)P(t=k|g_v(x))
\end{split}
\end{equation}

Where $t=k$ represents the $k_{th}$ expert is responsible for the input, $m$ is the number of views.

\section{Experiments and Evaluations}

We test our model on the Modified National Institute of Standards and Technology (MNIST) handwritten digits dataset. We also use the Canadian Institute For Advanced Research (CIFAR) 10/100 dataset, which is a collection of images. For a real-world scenario and to address some of the challenges in our healthcare monitoring research, we use the Technology Integrated Health Management (TIHM) dataset \cite{enshaeifar2018health}. The TIHM dataset consists of several sensor data types collected using in-home monitoring technologies from over 100 homes of people affected by dementia continuously for six months. The data includes environmental sensory data such as movement, home appliance use, doors open/closed, and physiological data such as body temperature, heart rate and blood pressure. The data was processed by a set of analytical algorithms to detect conditions such as hypertension, Urinary Tract Infections and changes in daily activities \cite{enshaeifar2019machine}. A clinical monitoring team used the results of the algorithms on a digital platform that we have developed in our previous work \cite{enshaeifar2018internet} and and whenever applicable verified the results for true or false positives. One of the key limitations of our previous work in TIHM was that the algorithms were trained offline, and they did not learn continually.  Another limitation was that by using conventional adaptive models, the algorithms changed over time when the environmental or health conditions; for example, due to seasonal effects or short-term illness. However, when an earlier learned status is re-observed by the models, the algorithms were not able to perform efficiently due to significant parameter changes. To evaluate the performance of our proposed continual learning and to demonstrate the effectiveness of the model in addressing a real-world problem, we evaluate TCNN, which is the Mv-TCNN without multi-view functions, on the TIHM dataset, which is not applicable with multi-view functions, and show it can address the challenges mentioned above. The other reason that for this time-series dataset, we use the model without multi-view function is that the data set has significantly lower dimensions compared to the image data in other datasets such as MNIST or CIFAR. Applying multi-view functions on this lower dimensionality data would create highly correlated augmented samples which in return will quickly overfit the model. The analysation of Mv-TCNN and TCNN is shown in ablation study. 
We pre-define the covariance matrix in the probabilistic layers as an identity matrix. Some of the multi-view functions modify the images randomly, e.g. the rotation functions rotate the images in a random angle. We feed the test data into multi-view functions ten times during the inference phase. The weight factor $\lambda$ in Equation (\ref{eq:loss_function}) is set to 0.1. The number of kernels in the probabilistic layer is pre-defined as the number of classes in each task.

We compare our model with several state-of-the-art approaches in different scenarios, as we mentioned in \ref{sec:reason}.
Based on the scenarios that are discussed in Section \ref{sec:reason}, we compare the proposed methods with: \textit{S1}, Continual Neural Dirichlet Process Mixture (CN-DPM) \cite{lee2020neural}. i) \textit{S2}: Incremental Moment Matching (IMM) \cite{lee2017overcoming}, Orthogonal Weight Modification (OWM) \cite{zeng2018continuous}, Reservoir \cite{chaudhry2019tiny} and Gradient episodic memory (GEM) \cite{lopez2017gradient}. The approaches of in the \textit{S3}, which informs the model of the task information in the test state, will increase the performance significantly \cite{van2019three}. However, the task information is usually unknown in practice. To avoid confusion, we do not compare to the methods which need the task information during the test state.

We also train two models as baselines. The offline represents a model which is trained with all the data at once. The online represents a model which is trained as a expandable model. More specifically, there will be multiple experts corresponding to different tasks. In the offline model, we optimise the model and hyper-parameters to obtain the best performance. In the online model, the architecture and the training configurations, including batch size, number of epochs are the same as the Mv-TCNN settings in each experiment.

\subsection{Split MNIST Experiment}
\label{sec:split_mnist_experiment}
The split MNIST experiment is a benchmark experiment in continual learning field \cite{zenke2017continual,lee2017overcoming,masse2018alleviating}. We split the MNIST to 5 different tasks of consecutive digits referring to 0/1, 2/3, 4/5, 6/7, 8/9. The basic architecture of an expert is a convolutional neural network with two convolutional and fully-connected layers. The pattern layer contains two data patterns.

As shown in Table \ref{tab:split_mnsit_compare}, the proposed model outperforms the state-of-the-art models in this experiment and similar to a model which is trained offline and after observing all the changes. 

\begin{table}[h!]
    \centering
    \caption{Average test accuracy on the split MNIST experiment.}
    \label{tab:split_mnsit_compare}
    \begin{tabular}{c|c|c}
    \hline
      \textbf{Approach} & \textbf{Method  }  & \textbf{Test Accuracy(\%)}  \\
     \hline
    \hline
    \multirow{2}{*}{Baseline} & offline &  97.94   \\
    &online & 44.90\\
    \hline
    \multirow{2}{*}{Regularisation} & OWM & 93.55  \\
     & IMM & 68.32  \\
    \hline
    \multirow{2}{*}{Memory Replay} & GEM & 92.20 \\  & Revisor & 85.69\\
    \hline
    \multirow{2}{*}{Dynamic Networks} & CN-DPM & 93.81 \\
    & \textbf{Mv-TCNN(proposed)} & \textbf{96.50} \\
    \hline
    \end{tabular}
\end{table}

\begin{figure*}[h!]
\centering
  \subfloat[First expert \label{fig:split_mnist_tlh_1}]{%
       \includegraphics[width=0.3\linewidth]{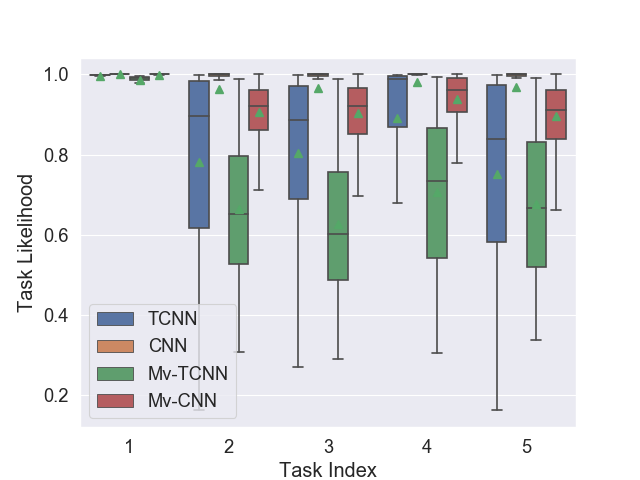}}
  \subfloat[Second expert \label{fig:split_mnist_tlh_2}]{%
       \includegraphics[width=0.3\linewidth]{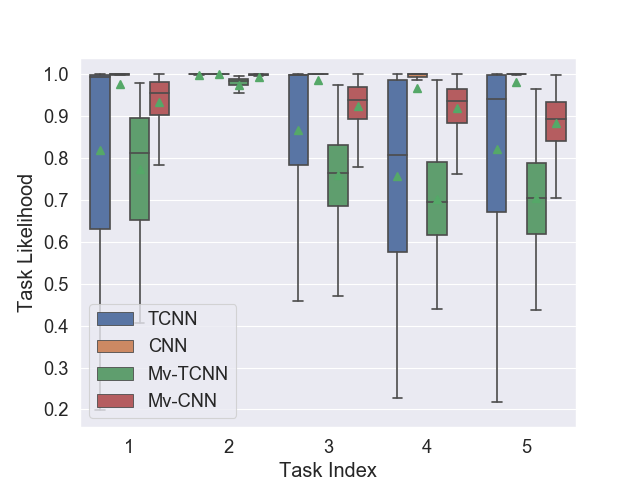}}
  \subfloat[Third expert \label{fig:split_mnist_tlh_3}]{%
       \includegraphics[width=0.3\linewidth]{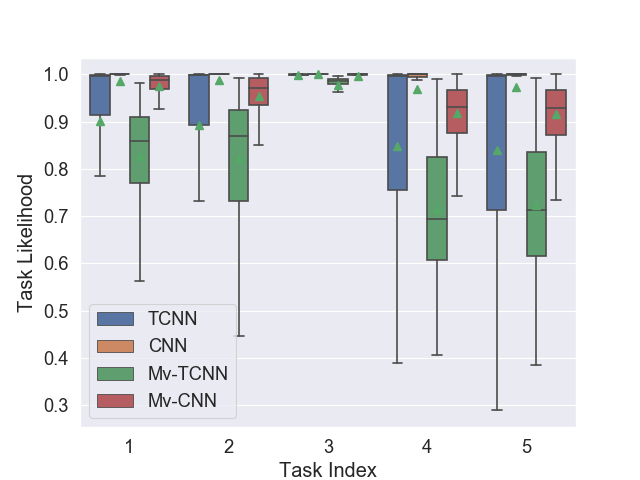}}\\
  \subfloat[Forth expert \label{fig:split_mnist_tlh_4}]{%
       \includegraphics[width=0.3\linewidth]{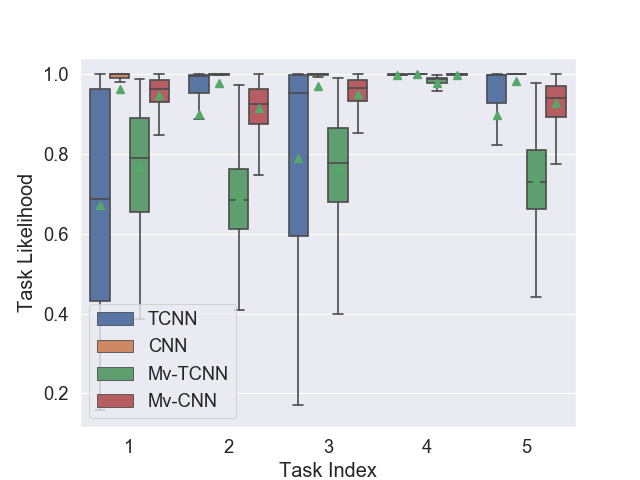}}
  \subfloat[Fifth expert \label{fig:split_mnist_tlh_5}]{%
       \includegraphics[width=0.3\linewidth]{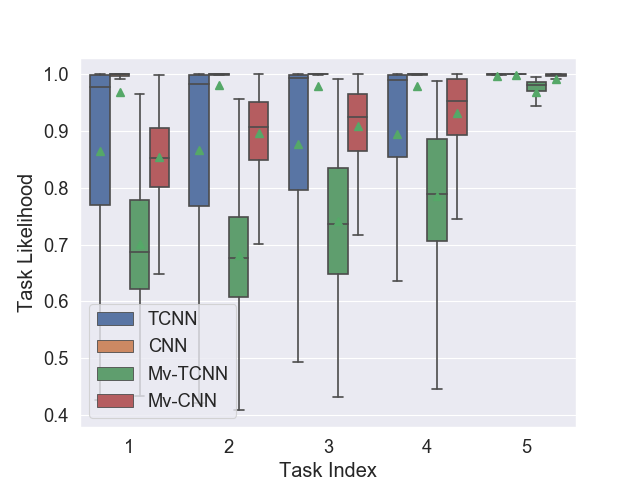}}
  \subfloat[Right selection rate \label{fig:split_mnist_tlh_tcnn}]{
       \includegraphics[width=0.3\linewidth]{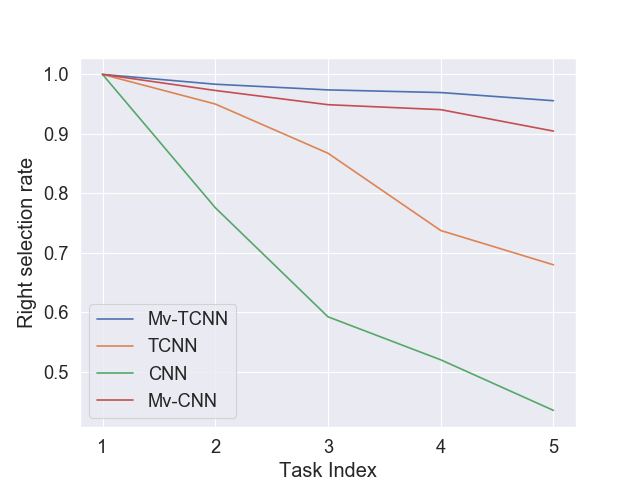}}
\caption{Ablation study of MNIST experiment. All the models are trained in expandable manner and have the same basic architecture. Figures \ref{fig:split_mnist_tlh_1} to \ref{fig:split_mnist_tlh_5} are the task Likelihood for each expert to each task. Figure \ref{fig:split_mnist_tlh_tcnn} is the rate of the model select the right expert to give the prediction. Overall, the likelihood of $k_{th}$ expert in Mv-TCNN on the $k_{th}$ tasks is distinguishable from task likelihood of the expert on other tasks.}
\label{fig:split_mnist_tlh}
\end{figure*}

\subsection{Split CIFAR10 Experiment}
\label{sec:split_cifar10_experiment}
In the second experiment, we increase the complexity of the dataset. We split CIFAR10 to 5 tasks in the same manner as MNIST experiment, and compare the performance of Mv-TCNN with the state-of-the-art methods. The based model is a convolutional neural network contains three convolutional layers and three fully-connected layers. The results is shown in Table \ref{tab:split_cifar10_compare}

\begin{table}[h!]
    \centering
    \caption{Average test accuracy on the split CIFAR10 experiment.}
    \label{tab:split_cifar10_compare}
    \begin{tabular}{c|c|c}
    \hline
      \textbf{Approach} & \textbf{Method  }  & \textbf{Test Accuracy(\%)}  \\
     \hline
    \hline
    \multirow{2}{*}{Baseline} & offline  &  93.17   \\
    & online & 45.35\\
    \hline
    \multirow{2}{*}{Regularisation} & OWM & 52.83  \\
     & IMM & 32.36  \\
    \hline
    Memory Replay  & Reservoir & 43.82\\
    \hline
    \multirow{2}{*}{Dynamic Networks} & CN-DPM & 46.98 \\
    & \textbf{Mv-TCNN(proposed)} & \textbf{57.36} \\
    \hline
    \end{tabular}
\end{table}

\subsection{split CIFAR100 experiment}

In the previous experiments, we only test the model to continual learn five different tasks. In this experiment, we continue to test the model in a more difficult scenario. We split the CIFAR100 datasets into 20 different tasks. Each of the tasks contains five different classes. The models have to perform a 100-way classification task. Overall, the complexity of each task and the number of tasks to be learned are significantly increased compared with the split CIFAR10 experiment. We have shown the result in the Table \ref{tab:split_cifar100_compare}. 

\begin{table}[h!]
    \centering
    \caption{Average test accuracy on the split CIFAR100 experiment.}
    \label{tab:split_cifar100_compare}
    \begin{tabular}{c|c|c}
    \hline
      \textbf{Approach} & \textbf{Method  }  & \textbf{Test Accuracy(\%)}  \\
     \hline
    \hline
    \multirow{2}{*}{Baseline} & offline &  73.08  \\
    & online  & 13.87\\
    \hline
    Memory Replay  & Reservoir & 10.01\\
    \hline
    \multirow{2}{*}{Dynamic Networks} & CN-DPM & 20.10 \\
    & \textbf{Mv-TCNN(proposed)} & \textbf{30.06} \\
    \hline
    \end{tabular}
\end{table}

\subsection{Healthcare Monitoring Data Experiment}
In this experiment, we would like to show that the proposed model can work for the data which cannot be augmented by the multi-view functions. The basic model we used in this experiment is Task Conditional Neural Network (TCNN), which is the same as Mv-TCNN but without augmenting the data using multi-view functions. 

For all the methods, we use a fully-connected neural network with two hidden layers as the underlying architecture. For the offline method, we use semi-supervised learning to optimise the results. The TIHM dataset is used in these experiments. This dataset is collected by obtaining consent from all the participants in the study and have received approval from an ethics review panel. Due to the personal and sensitive nature of the data and an ongoing trial, we cannot make the dataset publicly available. However, we have provided an anonymised snippet of the dataset to illustrate the features within the dataset\footnote{\url{https://github.com/mozzielol/Mv-TCNN/tree/master/datasets}}. 

We first evaluate changes in the daily activities of the participants in the study. This data contains three classes: low, medium and high levels of changes in the routine of daily living activities. Compared to the other experiments discussed above, this is a more challenging problem. The TIHM data is imbalanced. The low activity-change class contains 11057 samples; medium activity-change class includes 1146 samples, and high activity class contains only 64 samples. The model should be able to learn several tasks sequentially and also process the imbalanced data automatically. Different levels of activity and their changes also have various characteristics in different participants.  In other words, a change in the level of activities to indicate low or medium or high activity does not have the same distribution in all the participants' data. We split the dataset into training and test sets in the proportion of 9 to 1, and then follow the same steps as described above for other experiments to learn the three tasks in this experiment. Each task in this experiment only contains one class. The latter means that the test accuracy will be the same as decision accuracy. We have shown the evaluation results in the Table \ref{tab:tihm_test_accuracy}.

\begin{table}[h!]
    \centering
    \caption{Average test accuracy on the split TIHM activities experiment.}
    \label{tab:tihm_test_accuracy}
    \begin{tabular}{c|c|c}
    \hline
      \textbf{Approach} & \textbf{Method  }  & \textbf{Test Accuracy(\%)}  \\
     \hline
    \hline
    \multirow{2}{*}{Baseline} & offline &  98.77  \\
    & online  & 33.33\\
    \hline
    Regularisation  & IMM & 33.33\\
    \hline
    Dynamic Networks & \textbf{TCNN(proposed)} & \textbf{78.80} \\
    \hline
    \end{tabular}
\end{table}

We also evaluate the applicability of the model in classifying the cases of Urinary Tract Infections (UTIs) in the dataset. UTIs are one of the common causes of hospital admissions in people with dementia. In the TIHM dataset, we have some cases that are tagged by a monitoring team as true positives or false positives. The underlying data associated with these detected conditions are multivariate sensory data coming from sleep, movement, door and physiological monitoring sensors. One of the key limitations in our previous work in this area \cite{enshaeifar2019machine} was that the algorithms had to be trained offline and also they could not adapt to various distributions representing the patient groups that had UTIs with a different manifestation of symptoms. Using TCNN and RTCNN with the TIHM data, the model can incrementally learn different distributions in each class (i.e. positive or negative for UTIs) and to adapt to the changes in the input data.  The results of the test accuracy for this experiment are shown in Table \ref{tab:tihm_dementia}.

\begin{table}[h!]
    \centering
    \caption{Average test accuracy on the split TIHM activities experiment.}
    \label{tab:tihm_dementia}
    \begin{tabular}{c|c|c}
    \hline
      \textbf{Approach} & \textbf{Method  }  & \textbf{Test Accuracy(\%)}  \\
     \hline
    \hline
    \multirow{2}{*}{Baseline} & offline &  85.49  \\
    & online  & 50.00\\
    \hline
    Regularisation  & IMM & 50.00\\
    \hline
    Dynamic Networks & \textbf{TCNN(proposed)} & \textbf{70.48} \\
    \hline
    \end{tabular}
\end{table}

\section{Discussion}
In this section, we take the split MNIST experiment as an example to analyse the the overall performance of Mv-TCNN. We report the performance of Mv-TCNN under unknown task settings. We visualise the density approximated by the probabilistic layer. We also discuss the probabilistic layer and provide an ablation study.

\subsection{Revisit the learned tasks and learn new tasks}
\label{sec:task_unknown}
We analyse Mv-TCNN under the task unknown setup. The task-unknown settings represent conditions in which we do not provide the task information to the model at any time. We formulate the setting into two different phases:
i) Training Phase: the model learns a new task; ii) Prediction Phase: the model provides the results with test data and detects the changes. The model will go to the training phase if a change is detected in the prediction phase. In the prediction phase, we let the model detect the changes and give predictions on the learned tasks and and learning new tasks. Overall, the model should not detect the changes in the learned tasks and should be able to detect the changes in the new task.

\begin{figure}[h!]
\centering
  \subfloat[ \label{fig:mnist_con}]{%
       \includegraphics[width=\linewidth]{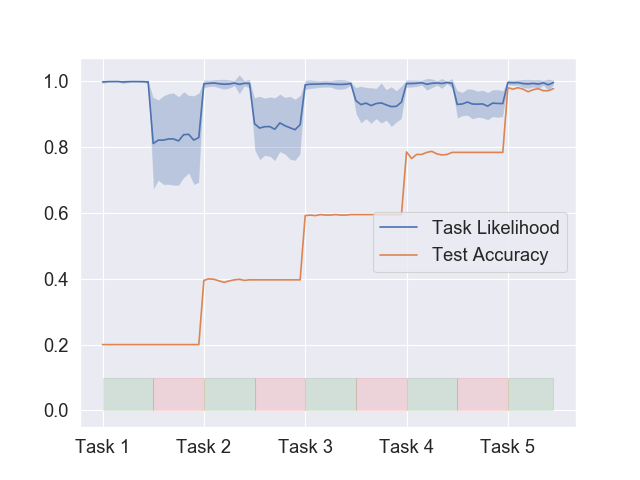}}\\
    \caption{MNIST experiment. Mv-TCNN learns five different tasks sequentially without having the task information in advance. The green blocks at the bottom represent the model tested on the learned task; the red blocks at the bottom represent the model tested on an unseen task. The blue shadows represent the area of $mean(tlh) \pm std(tlh)$, where $tlh$ is the task likelihood. When a change is detected, Mv-TCNN adapts to the new task automatically without forgetting the previous ones.}
    \label{fig:mnist_consistent}
\end{figure}

To detect the changes, we set a threshold which is the $mean - std$ of the task likelihood of the model on the current training samples. If the task likelihood on the test samples is less than the threshold in three consecutive batches, the change is detected and start to learn a new expert. We use three consecutive batches to avoid false positives in detecting task changes when there are transient changes in the data. The task-likelihood and the average test accuracy on all the tasks are shown in Figure \ref{fig:mnist_consistent}. Before and after adding a new expert for a new task, we test the model on ten batches, as shown below.

As shown in Figure \ref{fig:mnist_consistent}, while the model is tested on the learned task, the task likelihood remains on a higher-level. While a new task transpires, the task likelihood decreases significantly. The average test accuracy shows that Mv-TCNN detects the task changes and adapt to the new tasks quickly without forgetting the previously learned ones. Overall, Mv-TCNN provides a unique and novel feature by automatically detecting and adapting to new tasks in an efficient way.

\subsection{Analyse the Estimated Density}
\label{sec:dis_ayalyse}
In this section, we continue to use the split MNIST experiment to visualise the density estimated by the probabilistic layer.

\begin{figure}
\centering
  \subfloat[Sample from: Task 1, Data pattern from: expert 1\label{fig:diff_1}]{%
       \includegraphics[width=\linewidth]{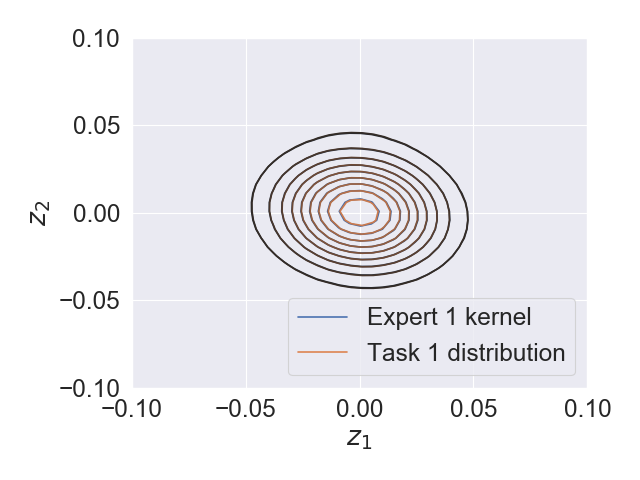}}\\
\centering
  \subfloat[Sample from: Task 2, Data pattern from: expert 1\label{fig:diff_2}]{%
       \includegraphics[width=\linewidth]{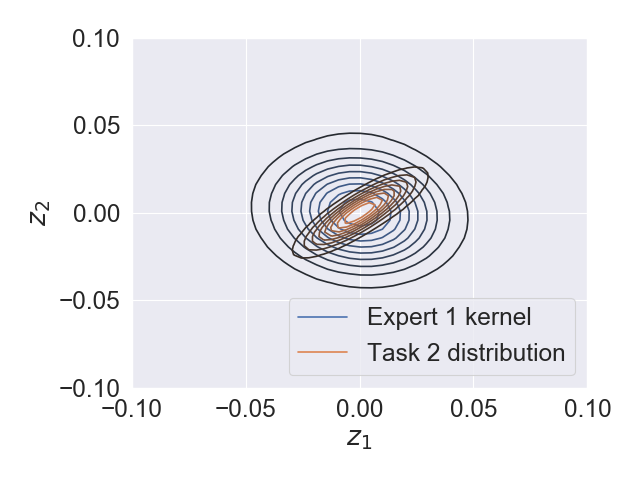}}
\caption{Difference between the samples distribution and the the task specific neurons in in the probabilistic layers. Sample are drawn from Task 1.}
\label{fig:distr_diff}
\end{figure}

As we mentioned in section \ref{sec:method_pnn}, the probabilistic layer can estimate the density of the inputs and remember the data patterns. The hidden layers can be viewed as a pattern extractor to map the inputs to the corresponding data patterns. If the expert observes the input beforehand, the data patterns extracted by the hidden layers should match the kernels in the probabilistic layer. In contrast, the extracted pattern cannot match the existing kernels in the probabilistic layer if the expert does not see the input beforehand. To verify this, we take the first expert in the split MNIST experiment and visualise the data patterns from the first and second tasks. The results are shown in Figure \ref{fig:distr_diff}. The distribution of the samples from task 2 is significantly different from the kernel in the expert 1. Hence the task likelihood will be decreased significantly.

\subsection{Ablation Study}
To do the ablation study, we build four different models: 1) Mv-TCNN: the proposed model consisting of multi-view functions, probabilistic layer. 2) TCNN: a variation of the proposed model that has a probabilistic layer but does not use multi-view functions. 3) Mv-CNN: a convolutional neural network. The model is trained with multi-view augmentation. 4) CNN: a convolutional neural network without any modification. All the models are trained in a generalisable manner. In other words, each model consists of several experts corresponding to different tasks. 

As shown in Figure \ref{fig:split_mnist_tlh}, the green triangles in the Figure \ref{fig:split_mnist_tlh_1} to \ref{fig:split_mnist_tlh_5} represent the mean of task likelihood. In the CNN, the triangles are all close to 1, and all the standard deviations of the task likelihood are quite small. This means the experts in the CNN tend to be confident with all the inputs even if the input does not belong to the experts. In the TCNN, the mean of the task likelihood is quite far from 1 if the input does not belong to the expert. However, the expert has a higher task likelihood for a portion of out-task samples. For example, the first expert has a relatively high task likelihood for the sample from task 4. The higher likelihood for the out-task samples will make the model assign the test data to a wrong expert and cannot give correct predictions. The standard deviation of the task likelihood is quite significant in the TCNN. In other words, the model does not have the most accurate way to predict the task likelihood for the out-task samples. In contrast to TCNN, Mv-CNN has a smaller standard deviation, but the mean values of the experts for the out-task samples is not small enough to distinguish with the in-task samples. In the Mv-TCNN, the task likelihoods for the in-task samples are easily distinguishable from the out-task samples. 

Figure \ref{fig:split_mnist_tlh_tcnn} shows the selection rate with respect to the number of tasks have learned. The selection rate indicates that the model selects the right expert to give a prediction on the test data. Both of the two proposed techniques (i.e. the multi-view and probabilistic layers) help the model to infer the task identity significantly better than the conventional method. The proposed Mv-TCNN model can infer the task identity efficiently and accurately compared with all the baseline models. .

\section{Conclusion}
In this paper, we first discuss the reasons for forgetting problem in machine learning when different tasks are given to a model at different times. We demonstrate how providing or acquiring the learning task information is essential to address this issue. We also present a challenge that we have in our healthcare monitoring research and discuss how an automated and scalable model can help to solve this issue in dynamic and changing environments. We then propose a Multi-view Task Conditional Neural Network (Mv-TCNN) model for continual learning of sequential tasks. Mv-TCNN is a novel expandable model that consists of multi-view functions to augment the data and and utilises a probabilistic layer to estimate the task likelihood. 

Mv-TCNN consists of several different experts corresponding to different tasks. The model can learn and decide which expert should be chosen and activated under different tasks that are given to the model, without having provided the task information in advance. Mv-TCNN can detect the changes in the tasks and learn new tasks automatically. The proposed model implements these features by using a probabilistic layer and measuring the task likelihood given a set of augmented samples associated with a specific task. 

Our proposed model outperforms the state-of-the-art methods in terms of accuracy and also performs well with both image and numerical time-series data. We have also evaluated Mv-TCNN using an imbalanced dataset. . We have shown how Mv-TCNN is used to identify the changes in the data and targets. We have also shown that the model can still perform well on the data which is not suitable for multi-view functions and use the previously learned parameters for each task to detect and predict changes in daily-living activities in our remote healthcare monitoring dataset. We hope this work can inspire more studies on investigating continual learning methods under dynamic environments.The future work will focus on ....

\bibliographystyle{IEEEtran}
\bibliography{ref}
%



%
\ifCLASSINFOpdf
\end{document}